# A General Neural Network Hardware Architecture on FPGA

Yufeng Hao
Dept. of Electronic, Electrical and
Systems Engineering
University of Birmingham, Edgbaston,
Birmingham, B152TE, UK
Email: yxh663@student.bham.ac.uk

*Abstract*—**Field Programmable Gate Arrays (FPGAs) plays an increasingly important role in data sampling and processing industries due to its highly parallel architecture, low power consumption, and flexibility in custom algorithms. Especially, in the artificial intelligence field, for training and implement the neural networks and machine learning algorithms, high energy efficiency hardware implement and massively parallel computing capacity are heavily demanded. Therefore, many global companies have applied FPGAs into AI and Machine learning fields such as autonomous driving and Automatic Spoken Language Recognition (Baidu) [1] [2] and Bing search (Microsoft) [3]. Considering the FPGAs great potential in these fields, we tend to implement a general neural network hardware architecture on XILINX ZU9CG System On Chip (SOC) platform [4], which contains abundant hardware resource and powerful processing capacity. The general neural network architecture on the FPGA SOC platform can perform forward and backward algorithms in deep neural networks (DNN) with high performance and easily be adjusted according to the type and scale of the neural networks.**

1 Introduction

The basic elements of FPGA are Configurable Logic Block (CLB) and interconnecting resources [5], which are flexible and convenient to implement interfaces such as I2C, SPI, and so on, and control circuits for specific requirements since FPGA was firstly introduced about 30 years ago. And with the development of integrated circuit (IC) technology, FPGA has held an outstanding performance in low-power and large-scale parallel computing domain. Compared to CPU or GPU which are based on Von Neumann or Harvard Architecture, FPGA has a more flexible framework to implement algorithms. The instructions and data in FPGA can be designed in a more efficient way without constraint of fixed architectures, which is suitable for designers to explore the high performance implement approaches in power or computing sensitive fields. Besides, though its clock frequency is slower than that of CPU or GPU, FPGA usually executes operations in a few clock periods, which makes FPGA to hold a competitive advantage in real-time data processing and low power consumption design, considering CPU and GPU need dozens of instructions to execute one operation and higher frequency leads to higher power consumption. Meanwhile, considering its inherent parallel architecture, FPGA shows a powerful processing capacity in massive convolution, multiply-accumulation, and other matrix operations which are essential in current neural network or machine learning algorithms. Therefore, it is indispensable to apply FPGA into the above fields to gain cost and real-time computing advantage.

## 2 Objective

We will build a general neural network hardware architecture on FPGA, which has an outperformance in energy efficiency and real-time computation. Based on the architecture, different types and scales of neural networks can be implemented and the neural network training and deployment can be directly performed on FPGA.

## 3 Hardware Architecture Implement

### 3.1 Neural Networks Perspective

Artificial Neural Network consists of neuron cells, which are connected to each other and arranged in layers [6]. Basically, one neural network has only one input layer and one output layer but can hold many hidden layers. Each layer is composed of neuron cells. Each neuron cell can be regarded as a nonlinear transformation unit which holds different weights as the multiplier for different input data from the previous layer. The basic unit of a neuron cell can be illustrated as following.

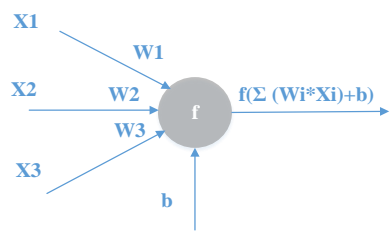

Figure 3-1 neuron cell unit

The system diagram of General Neural Network is shown below as figure 3-2. (The picture is form http://neuralnetworksanddeeplearning.com/chap5.html )

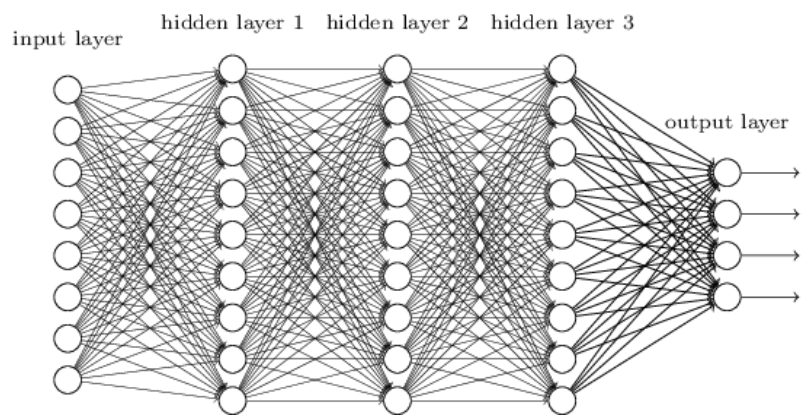

Figure 3-2 general neural network system diagram

According to the number of hidden layers, the neural network can be classified as Deep Neural Network (DNN) which usually has more than 2 layers and shallow neural network. In terms of the direction of data flow, there are Recurrent Neural Network (RNN) where there is a data flow between adjacent cells in the same layer and general Neural Network where there is not such connection. Judged from the operations executed by the neural network, there is a Convolutional Neural Network (CNN) which mainly perform its function through convolutional operation. The nonlinear transformation layers mentioned above are the core elements for all types of neural network such as CNN, DNN, RNN, and so on. Essentially, it is the nonlinear transformation that the neural network possesses the capacity of extracting high dimension features of original data set.

Actually, Neural Network is a kind of data modeling algorithm with multi-layer and nonlinear transformation [7]. There are two main procedures in design neural networks: forward process and backward process [8]. In the forward process, we define the neural network architecture such as how many layers, how many cells in each layer, and which kinds of cells we choose, and then the input data can flow to the output. In the back process, we define the loss function to calculate the gap between the prediction value and the label values. Through calculating the gradient in the backward path, we upgrade the weights of the neural network. One forward and one backward process can be called as one epoch, and it is the training of neural network. Usually, a trained neural network model need hundreds of epochs to be available.

### 3.2 General Neural Network Architecture implement

According to the general neural network structure which is shown in figure 3-2, we implement general framework of a general neural network on FPGA shown in Figure 3-3, which contains the forward propagation, backward propagation, and control modules to complete the training of the neural network. In the framework, we use the highly reusable modules to perform the neural network matrix operations, which makes it easy to adjust the types or the number of hidden layers of a neural network.

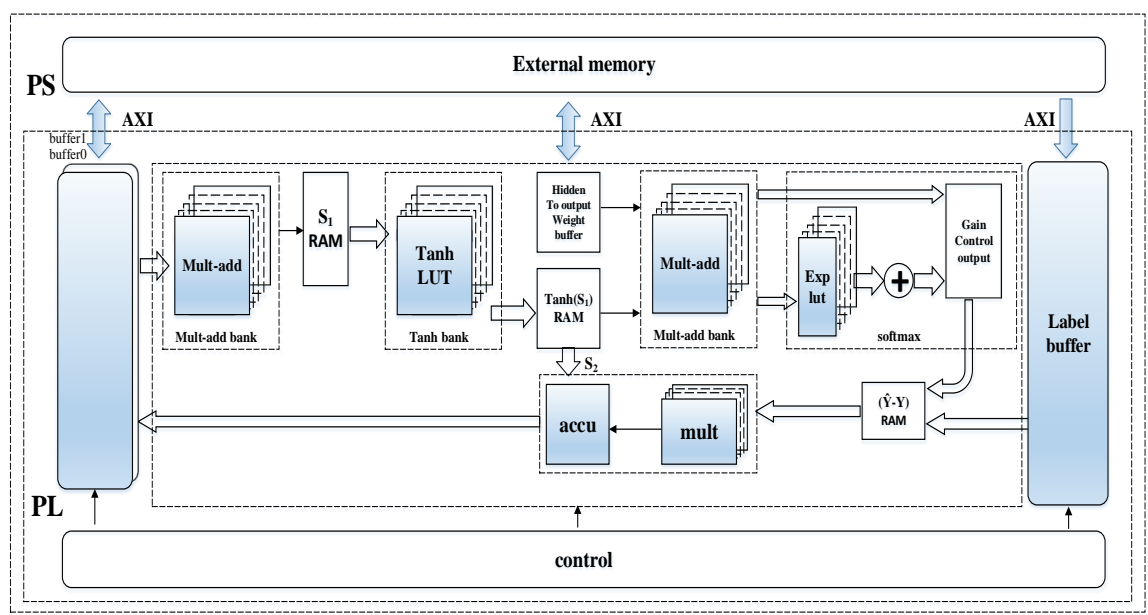

Figure 3-3 general neural network hardware architecture on FPGA SOC

The general neural network hardware architecture framework is shown as the figure 3-3. The framework includes two processes: the forward process which generates the prediction values and the backward process which upgrades the weights according to the loss values. The mult-add bank, RAM, and Tanh bank modules can perform one layer forward propagation operations. They can be reused to adjust the neural network or increase the number of hidden layers. The accu and mult modules will perform the back propagation operations. We will give a more detail explanation about the hardware implement framework in the following paragraphs.

For the forward process, there are 3 matrix operation stages: input layer to hidden layer, hidden layer to hidden layer, and hidden layer to output layer. **In the first stage**, the input vectors $X_i$ (i = 1,2,...T) multiply the first hidden layer weights matrix $W^{H1XT}$ (H1 equals to the dimension of the first hidden layer, T equals to the length of input vectors), which are loaded form the buffer0 and buffer1. We use mult-add bank to accomplish the above matrix operation. The mult-add bank consists of many parallel multiplication and accumulation units, whose number can be adjusted according to the dimension of hidden layer and logic resources on FPGA. Then the result matrix is stored in the $S_1$ RAM as the input of activation function. There are lots of activation functions which can be chose in practice such as sigmoid, Rectified Linear Unit (ReLU), Tanh, and so on. We implement the activation functions by Lookup Table (LUT) through which different initialization parameters can be loaded in order to perform different activation functions. The output matrix of the first hidden layer (M1) is stored in the Tanh($S_1$) RAM. **In the second stage**, the M1 multiplies the second hidden layer weights matrix $W^{H1XH2}$ (H2 equals to the dimension of the second hidden layer), which still is implemented by mult-add bank. In figure 3-3, we only demonstrate a general neural network with one hidden layer. The multi-hidden layers neural network hardware implement can be carried out through the reuse of the mult-add bank, weights RAM, and tanh bank units. **The final stage** in the forward process is to perform the matrix mapping from last hidden layer to the output layer. According to the expression of softmax function: $\sigma(z_j) = \frac{e^{z_j}}{\sum_{k=1}^{k} e^{z_k}}$ , we can similarly implement the exponential function by LUT, then get the sum by addition units. According to experience value, we set a maximum limit about the sum (eg. 1024) and then adjust the output as the ratio of sum and the maximum limit.

For the backward process, we use cross-entropy loss function [9] [10] to calculate error derivatives in back propagation algorithm. The detail of back propagation algorithm operation is described in [11]. The core step is to calculate the multiplication of the error and the derivative of the activation function. In implementation level, the back propagation can be described as: $\gamma \sum_{i=1}^{m} (\hat{y}_i - y_i) \otimes S_2$ . The $(\hat{y}_i - y_i)$ is the disparity of the prediction value and the label. The $S_2$ is the input vectors to the layer where the residual signals generates in the back propagation path. The $\otimes$ symbol means exterior product operation, which can be performed by mult module in the framework.

The control unit in the bottom of the diagram controls the neural network training processes, which provides the timing to read or write weights buffers, perform matrix operations, and store calculation results.

**4 Conclusion**

The whole neural network algorithms is implemented on the XILINX ZU9CG FPGA SOC platform, which contains abundant computing memory resources (2520 DSPs and 32Mb on-chip memory) and remarkable programming ability (Dual-core ARM Cortex A53). In the hardware framework, different deep neural networks can be implemented by reusing the forward propagation modules and slightly modifying the backward propagation modules. For a larger scale deep neural network, a cluster of FPGA can be integrated into a whole platform to perform the algorithm. We can also deploy deep learning framework tools such as TensorFLow into this 64-bit FPGA SOC platform, calling the FPGA hardware resources directly. This will provide a real-time, high energy efficiency, and fast deployment embedded solution for deep learning and machine learning applications.